\pdfoutput=1
\documentclass[letterpaper, 10 pt, conference]{ieeeconf}  

\IEEEoverridecommandlockouts                              
\usepackage[backend=biber,
            url=false,
            isbn=false,
            doi=false,
            backref=false,
            style=ieee,
            natbib=true,
            mincitenames=1,
            maxcitenames=1,
            citestyle=numeric-comp,
            sorting=nyt,
            block=none]{biblatex}
        
\renewcommand{\bibfont}{\small}
\usepackage{graphics}
\usepackage[pdftex]{graphicx}
\usepackage{wrapfig}
\DeclareGraphicsExtensions{.pdf,.png,.jpg}
\usepackage{epsfig}
\usepackage[font={small}]{caption}
\usepackage{subcaption}
\usepackage[rightcaption]{sidecap}
\usepackage{pbox}

\usepackage{bigstrut}
\usepackage{mathtools}
\usepackage{amsmath, amssymb, amscd, amsthm}
\usepackage{ wasysym } 
\usepackage{amsfonts}
\usepackage{mathptmx} 
\usepackage{gensymb} 
\usepackage{nicefrac}       
\numberwithin{equation}{section} 

\DeclareMathAlphabet{\mathcal}{OMS}{lmsy}{m}{n}
\DeclareSymbolFont{largesymbols}{OMX}{cmex}{m}{n}
\usepackage{textcomp} 
\usepackage{dsfont}

\usepackage{algorithm} 
\usepackage{algorithmic}

\usepackage{array} 
\usepackage{tabu}
\usepackage{multirow}
\usepackage{multicol}
\usepackage{booktabs}

\usepackage[T1]{fontenc} 
\usepackage[utf8]{inputenc}
\usepackage[english]{babel} 
\usepackage{units}
\usepackage{bm}
\usepackage{times} 
\usepackage{xspace}
\usepackage{balance} 
\usepackage{csquotes}
\usepackage{makeidx}
\usepackage{blindtext}

\let\labelindent\relax
\usepackage{enumitem}

\usepackage{ragged2e}
\usepackage{soul} 
\usepackage{subfiles} 

\usepackage[protrusion=true,expansion=true]{microtype}
\usepackage[yyyymmdd]{datetime}

\date{\protect\formatdate{1}{1}{2001}}

\usepackage{url}
\makeatletter
\g@addto@macro{\UrlBreaks}{\UrlOrds}
\makeatother
\usepackage{xcolor}

\newcommand{\tocite}[1]{%
\textcolor{red}{[cite:\ifthenelse{\equal{#1}{}}{}{#1}?]}
}

\newcommand{\ignore}[1]{}

\renewcommand{\baselinestretch}{1.0}



\newcommand{\algname}{Thompson Sampling with Learned Priors (TSLP)\xspace}
\newcommand{\algabbr}{TSLP\xspace}

\DeclareMathOperator{\argmax}{argmax}

\title{\LARGE \bf
Accelerating Grasp Exploration by Leveraging Learned Priors%
}

\author{Han Yu Li$^{*}$, Michael Danielczuk$^{*}$, Ashwin Balakrishna$^{*}$, Vishal Satish$$, Ken Goldberg$$%
\thanks{$^{*}$ equal contribution}%
\thanks{The AUTOLAB at UC Berkeley} \thanks{$\lbrace$katherine.li,mdanielczuk,ashwin\_balakrishna,vsatish,goldberg$\rbrace$@berkeley.edu}
}

\begin{document}

\maketitle

\begin{abstract}
The ability of robots to grasp novel objects has industry applications in e-commerce order fulfillment and home service. Data-driven grasping policies have achieved success in learning general strategies for grasping arbitrary objects. However, these approaches can fail to grasp objects which have complex geometry or are significantly outside of the training distribution. We present a Thompson sampling algorithm that learns to grasp a given object with unknown geometry using online experience. The algorithm leverages learned priors from the Dexterity Network robot grasp planner to guide grasp exploration and provide probabilistic estimates of grasp success for each stable pose of the novel object. We find that seeding the policy with the Dex-Net prior allows it to more efficiently find robust grasps on these objects. Experiments suggest that the best learned policy attains an average total reward $64.5\%$ higher than a greedy baseline and achieves within $5.7\%$ of an oracle baseline when evaluated over $300,000$ training runs across a set of $3000$ object poses.
\end{abstract}

\section{Introduction}
\label{sec:introduction}
Robotic grasping has a wide range of industry applications such as warehouse order fulfillment, manufacturing, and assistive robotics. However, grasping is a difficult problem due to uncertainty in sensing and control, and there has been significant prior work on both analytical~\cite{rimon2019mechanics,bicchi2000robotic,prattichizzo2016grasping,murray2017mathematical, wang2019adversarial} and data-driven methods~\cite{lenz2015deep,kappler2015leveraging,levine2018learning} for tackling these challenges. Recently, data-driven grasping algorithms have shown impressive success in learning grasping policies which generalize across a wide range of objects~\cite{mahler2019learning,james2019sim,morrison2018closing}. However, these techniques can fail to generalize to novel objects that are significantly different from those seen during training. Precisely, we investigate learning grasping policies for objects where general purpose grasping systems such as~\cite{mahler2019learning} produce relatively inaccurate grasp quality estimates, resulting in persistent failures during policy execution.

This motivates algorithms which can efficiently learn from on-policy experience by repeatedly attempting grasps on a new object and leveraging grasp outcomes to adjust the sampling distribution. Deep reinforcement learning has been a popular approach for online learning of grasping policies from raw visual
input~\cite{pinto2016supersizing,levine2018learning,kalashnikov2018qt}, but these approaches often take prohibitively long to learn robust grasping policies. These approaches typically attempt to learn \textit{tabula rasa}, limiting learning efficiency. In this work, we introduce a method which leverages information from a general purpose grasping system to provide a prior for the learned policy while using geometric structure to inform online grasp exploration. 
We cast grasp exploration in the multi-armed bandits framework as in~\cite{laskey2015multi, mahler2016dex}.
However, unlike ~\citet{laskey2015multi} which focuses on grasping 2D objects where some rough geometric knowledge is known and \citet{mahler2016dex} which presents a method to transfer grasps to new 3D objects using a dataset of grasps on 3D objects with known geometry, we focus on efficiently learning grasping policies for 3D objects directly from depth image observations. In addition, the algorithm learns to grasp a specific object through online interaction, unlike \citet{mahler2016dex} which learns a general grasping policy for arbitrary objects. Specifically, we present a method which leverages prior grasp success probabilities from the state-of-the-art Dex-Net 4.0 grasp quality network GQ-CNN~\cite{mahler2019learning} to guide online grasp exploration on unknown 3D objects with only depth-image observations. 

\begin{figure}[t!]
    \centering
    \includegraphics[width=\linewidth]{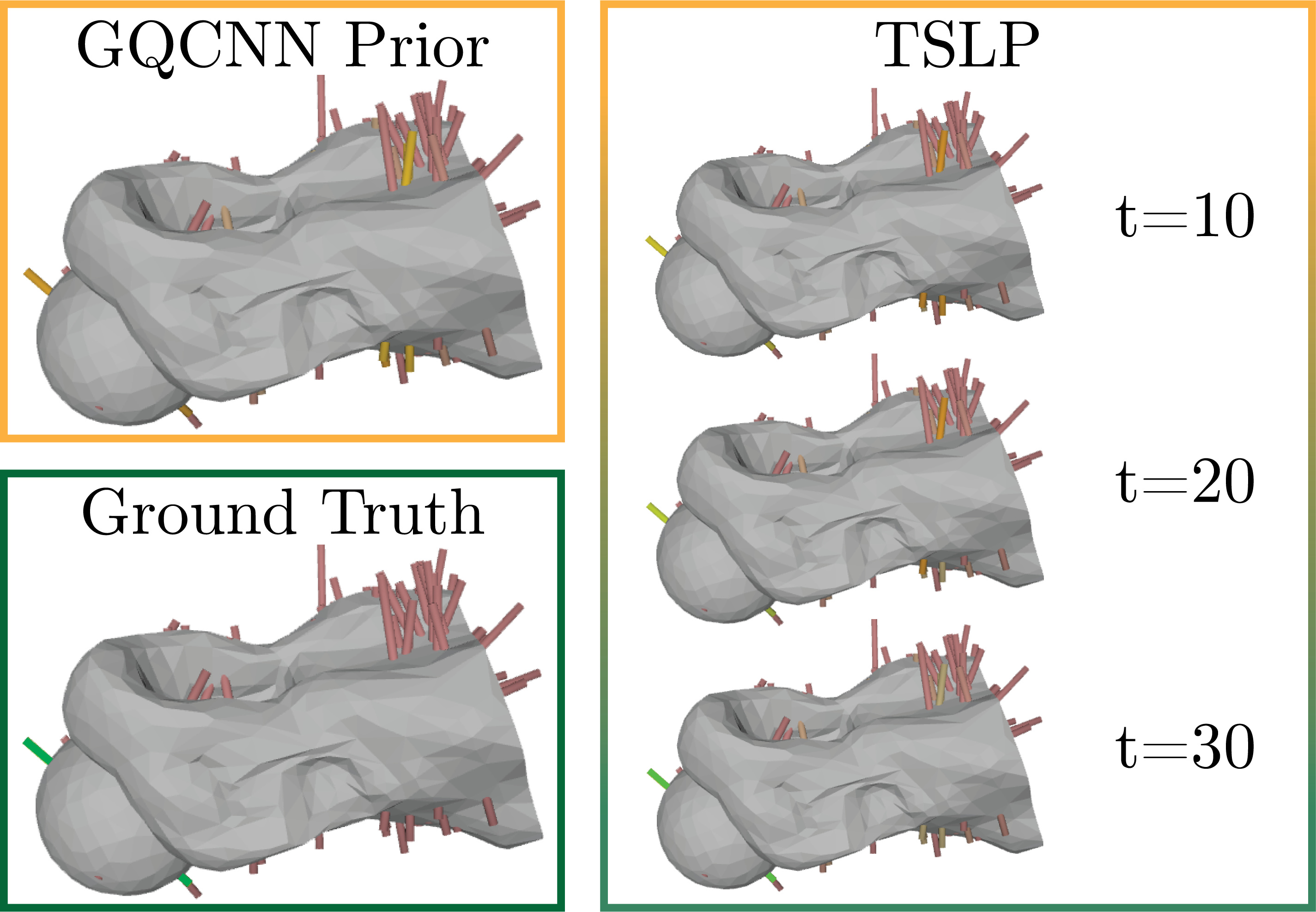}
    \caption{For adversarial objects, state-of-the-art grasp planning algorithms may incorrectly predict the distribution over grasp qualities (left column), where each whisker represents a grasp candidate colored by the likelihood of success (red indicates a poor grasp, green indicates a robust grasp). We find that \algabbr can use the prior to efficiently discover the best grasp on the object (right column). Here, the policy discovers the only robust grasp despite a poor initial estimate of its quality from the GQ-CNN prior.}
    \label{fig:splash}
\end{figure}

The contributions of this paper are:
\begin{enumerate}
    \item A new problem formulation for leveraging learned priors on grasp quality to accelerate online grasp exploration.
    \item An efficient algorithm, \algname, for learning grasping policies on novel 3D objects from depth images by leveraging priors from the Dex-Net 4.0 robot grasping system~\cite{mahler2017dex}.
    \item A new formulation of the mismatch between a prior distribution on grasp qualities and the ground truth grasp quality distribution and empirical analysis studying the effect of this mismatch on policy performance.
    \item Simulation experiments suggesting that \algabbr attains an average total reward $64.5\%$ higher than a greedy baseline when evaluated over $300,000$ training runs across $3000$ object poses and is able to effectively leverage information from a GQ-CNN prior.
\end{enumerate}
\section{Related Work}
Robot grasping methods develop policies that execute grasps on novel objects, and can be divided into analytical methods and data-driven methods. Analytic methods assume knowledge of the geometry of the object to be grasped~\cite{prattichizzo2016grasping,murray2017mathematical,bicchi2000robotic,rimon2019mechanics,Liu2020DeepDG} or use geometric similarities between known and unknown objects to infer grasps on unknown objects~\cite{mahler2016dex}
However, the generalization of these methods is limited for objects dissimilar to the known objects, or when geometric information is unknown~\cite{bohg2013data}, as in the case we consider. Data-driven methods rely on labels from humans~\cite{lenz2015deep,saxena2008robotic,kappler2015leveraging,morrison2018closing}, self-supervision across many physical trials~\cite{pinto2016supersizing,levine2018learning,kalashnikov2018qt,Bodnar2019QuantileQF}, simulated grasp attempts~\cite{johns2016deep,viereck2017learning}, or sim-to-real transfer methods such as domain randomization~\cite{bousmalis2018using} or domain adaptation~\cite{james2019sim}. Hybrid approaches generate simulated grasp labels using analytical grasp metrics such as force closure or wrench resistance~\cite{mahler2017dex,mahler2018dex,mahler2019learning}. These data-driven and hybrid approaches train a deep neural network  on the labeled data to predict grasp quality or directly plan reliable grasps on novel objects. A recent paper in sim-to-real transfer learning correct for inaccurate gripper poses predicted by the neural network by combining domain adaptation and visual servoing in the grasp planning process \cite{pedersen:hal-02495837}. However, for adversarial objects~\cite{wang2019adversarial}, for which very few high quality grasps exist, or for objects significantly out of the training distribution, grasps planned by these methods may still fail. The presented method aims to leverage learned grasp quality estimates to enable efficient online learning for difficult-to-grasp objects through physical exploration of one pose of one object at a time, without previous knowledge of the object's geometry.

Past works have formulated grasp planning as a Multi-Armed Bandit problem for grasping 2D objects where some geometric knowledge is known~\cite{laskey2015multi} or for transferring grasps to unknown 3D objects using a dataset of grasps on 3D objects with known geometry. \citet{laskey2015multi} found that Thompson sampling with a uniform prior significantly outperformed uniform allocation or iterative pruning in 2D grasp planning in terms of convergence rate to within 3\% of the optimal grasp, but their policy is limited to 2D grasps and cannot operate directly on visual inputs. \citet{mahler2016dex} extend~\cite{laskey2015multi} to 3D and incorporate prior information from Dex-Net 1.0, a dataset of over 10,000 3D object models and a set of associated robust grasps. The algorithm then uses Thompson sampling, in which the prior belief distribution for each grasp is calculated based on its similarity to grasps and objects from the Dex-Net 1.0 database~\cite{mahler2016dex}. For objects with geometrically similar neighbors in Dex-Net 1.0, the algorithm converges to the optimal grasp approximately 2 times faster than Thompson sampling without priors~\cite{mahler2016dex}. In contrast, we present a Bayesian multi-armed bandit algorithm for robotic grasping with depth image inputs that does not require a database to compute priors but instead leverages the Dex-Net 4.0 grasping system from~\cite{mahler2019learning} as a learned prior to guide active grasp exploration. Instead of learning a general grasping strategy for arbitrary objects as \cite{mahler2016dex}, the algorithm learns to grasp a specific object through online interactions with the object.

\section{Problem Statement}
\label{sec:prob-statement}
Given a single unknown object on a planar workspace, the objective is to effectively leverage prior estimates on grasp qualities to learn a grasping policy that maximizes the likelihood of grasp success. We first define the parameters and assumptions on the environment (Sections~\ref{subsec:assumptions} and~\ref{subsec:definitions}), cast grasp exploration in the Bayesian bandits framework (Section~\ref{subsec:bayesian-bandits}), and formally define the policy learning objective (Section~\ref{subsec:learning-obj}). 
\subsection{Assumptions}
\label{subsec:assumptions}
We make the following assumptions about the environment.
\begin{enumerate}
    \item \textbf{Pose Consistency: }We assume that the object remains in the same pose during all rounds of learning. In simulation, this can be achieved by using ground-truth knowledge of physics and object geometry. In physical experiments, the pose consistency assumption will not hold generally. We discuss methods to approximately enforce pose consistency in physical experiments in Section~\ref{sec:future-work}.
    \item \textbf{Evaluating Grasp Success: }We assume that the robot can evaluate whether a grasp has succeeded. In simulation, grasp success can be computed by using ground-truth knowledge of physics and object geometry. In physical experiments, success or failure can be determined using load cells, as in~\cite{mahler2019learning}.
\end{enumerate}

\subsection{Definitions}
\label{subsec:definitions}
\begin{enumerate}
    \item \textbf{Observation:} An overhead depth image observation of the object at time $t = 0$ before policy learning has begun, given by $o \in \mathds{R}_+^{H \times W}$. 
    \item \textbf{Arms:} We define a set of $K$ arms, $\lbrace a_k \rbrace_{k=1}^{K}$.
    \item \textbf{Actions:} Given a selected arm $k$ we define a corresponding grasp action $u_k \in \mathcal{U}$.
    \item \textbf{Reward Function:} Rewards for each arm are drawn from a Bernoulli distribution with unknown parameter $p_k$: $r(u_k) \sim \textrm{Ber}(p_k)$. Here $r(u_k) = 1$ if executing $u_k$ results in the object being successfully grasped, and $0$ otherwise.
    \item \textbf{Priors:} We assume access to priors on the Bernoulli parameter $p_k$ for each arm $k$.
    \item \textbf{Policy:} Let $\pi_\theta(u_k)$ denote a policy parameterized by $\theta$ which selects an arm $k$ and executes the action $u_k$. Thus, $\pi_\theta(u_k)$ defines a distribution over $\mathcal{U}$ at any given timestep $t$.
\end{enumerate}

\subsection{Bayesian Bandits}
\label{subsec:bayesian-bandits}
A multi-armed bandits problem is defined by an agent which must make a decision at each timestep $t \in \{1, 2, \hdots T\}$ by selecting an arm $k \in \left\lbrace 1,2, \hdots  K\right\rbrace$ to pull. After each arm pull, the agent receives a reward which is sampled from an unknown reward distribution. In the Bayesian bandits framework~\cite{TS-survey}, the agent maintains a belief over the parameters of the reward distribution for each arm, which can optionally be seeded with a known prior. The objective is to learn a policy with a distribution over arms that maximizes the cumulative expected reward over $T$ rounds.

\begin{figure*}[th!]
\centering
\vspace{4pt}
\includegraphics[width=\linewidth]{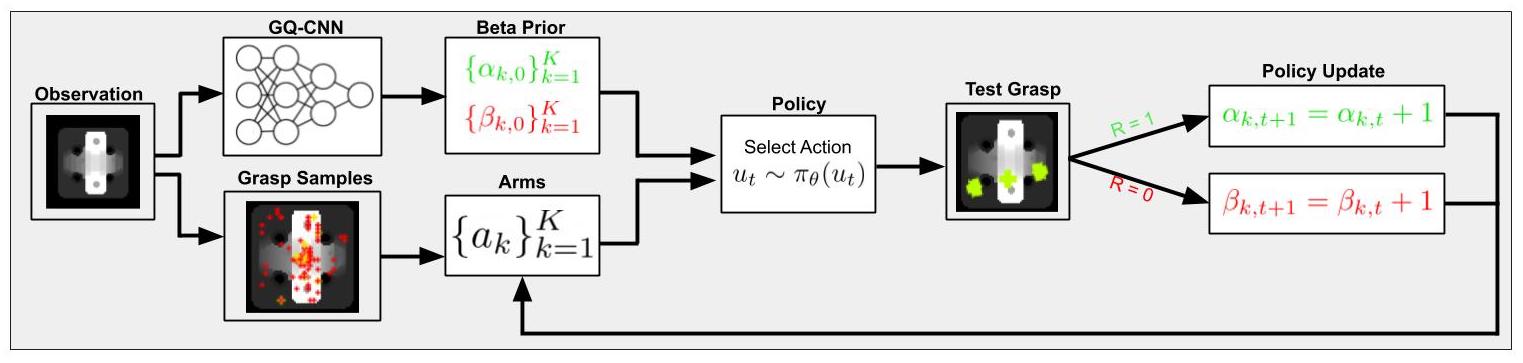}
\caption{\textbf{Method Overview: } A pre-trained GQ-CNN is used to set the priors on the reward parameters for each arm given the initial observation $o$ and arms are sampled on observation $o$. Then, at each timestep the learned policy selects an arm and executes the corresponding action in the environment. The Thompson sampling parameters are updated based on the reward received as described in Section~\ref{subsec:thompson-sampling}.}
\label{fig:system}
\end{figure*}

\subsection{Learning Objective}
\label{subsec:learning-obj}
The objective in policy learning is to maximize the total accumulated reward, which corresponds to maximizing the frequency with which the object is grasped. Let $u_{t}$ denote the action selected at timestep $t$. Then the objective is to learn policy parameters $\theta$ to maximize the following:
  \begin{align}
    \label{eq:objective}
    J(\theta) =  \mathds{E}_{u_t \sim \pi_\theta(u_t)} \left[  \sum_{t=1}^{T} r(u_t) \right]
\end{align}
\section{Grasp Exploration Method}
\label{sec:methods}
We discuss how to leverage learned priors from GQ-CNN to guide grasp exploration by using Thompson sampling, to learn a vision-based grasping policy. Since rewards are drawn from a Bernoulli distribution as defined in Section~\ref{sec:prob-statement}, we represent the prior with a Beta distribution, the conjugate prior for a Bernoulli distribution. As noted in~\cite{laskey2015multi}, this choice of prior is convenient since we can update the belief distribution over an arm $k$ after executing corresponding action $u_k$ in closed form given the sampled reward. See Figure~\ref{fig:system} for a full method overview.

\subsection{Thompson Sampling with a Beta-Bernoulli Process}
\label{subsec:thompson-sampling}
Given that we pull arm $k$ at time $t$ and receive reward $r(u_k) \in \{0, 1\}$, as shown in ~\cite{laskey2015multi}, we can form the posterior of the Beta distribution by updating the shape parameters $\alpha_{k, t}$ and $\beta_{k, t}$:
\begin{align*}
  \alpha_{k, t+1} &= \alpha_{k, t} + r(u_k)  \\
  \beta_{k, t+1} &= \beta_{k, t} + (1 - r(u_k))
\end{align*}

For Thompson sampling, at time $t$, the policy samples $\hat{p}_{k, t} \sim \text{Beta}(\alpha_{k, t}, \beta_{k, t})$ for all arms $k \in \{1, 2, \hdots K\}$, selects arm $k^* = \argmax_{k} \hat{p}_{k, t}$, and executes the corresponding action $u_{k^*}$ in the environment. Note that the expected Bernoulli parameter for arm $k$ can be computed from the current shape parameters $\alpha_{k, t}$ and $\beta_{k, t}$ as follows:
\begin{align}
  \label{eq:est_grasp_success_prob}
   \mathds{E}\left[\hat{p}_{k, t}\right] = \frac{\alpha_{k, t}}{\alpha_{k, t} + \beta_{k, t} }
\end{align}

However, it remains to appropriately initialize $\alpha_{k, 0}$ and $\beta_{k, 0}$. Note that setting $\alpha_{k, 0} = \beta_{k, 0} = 1 \ \forall \ k \in \{1, 2, \hdots K\}$ corresponds to a prior which is uniform on $[0, 1]$ for Bernoulli parameter ${p}_{k, t}$. We instead set $\alpha_{k, 0}, \beta_{k, 0}$ according to a learned prior by using the initial depth image observation $o$.

\subsection{Leveraging Neural Network Priors}
\label{subsec:NN-priors}
We use a pre-trained Grasp Quality Convolutional Neural Network (GQ-CNN) from~\cite{mahler2019learning} to obtain an initial estimate of the probability of grasp success. GQ-CNN learns a $Q$-function, $Q_\phi(\cdot, \cdot)$, which given an overhead depth image of an object and a proposed parallel jaw grasp, estimates the probability of grasp success. However, as explored in~\cite{wang2019adversarial}, there exist many objects for which the analytical methods used for training GQ-CNN are relatively inaccurate, resulting in significant errors. Thus, we refine the initial GQ-CNN grasp quality estimates with online exploration.

We first compute $Q_\phi(o, u_k) \ \forall \ k \in \{1, 2, \hdots K\}$ and use these estimates as each arm's initial mean Bernoulli parameter. Note that $\alpha_{k, t}$ and $\beta_{k, t}$, as defined in Section~\ref{subsec:thompson-sampling}, correspond to the cumulative number of grasp successes and grasp failures respectively for action $u_k$ up to time $t$. Thus, $(\alpha_{k, 0}, \beta_{k, 0})$ can be interpreted as pseudo-counts of grasp successes and failures respectively for action $u_k$ before policy learning has begun, while prior strength $S = \alpha_{k, 0} + \beta_{k, 0}$ can be interpreted as the number of pseudo-rounds before policy learning. If $S$ is large, the prior induced by $(\alpha_{k, 0}, \beta_{k, 0})$ will significantly influence the expected Bernoulli parameter given in~\ref{eq:est_grasp_success_prob} for many rounds, while if $S$ is small, the resulting prior will be quickly washed out by samples from online exploration. We enforce the following initial conditions for $(\alpha_{k, 0}, \beta_{k, 0})$, given the GQ-CNN prior:
\begin{align*}
    \frac{\alpha_{k, 0}}{\alpha_{k, 0} + \beta_{k, 0}} &= Q_\phi(o, u_k) \\
    \frac{\beta_{k, 0}}{\alpha_{k, 0} + \beta_{k, 0}} &= 1 - Q_\phi(o, u_k)
\end{align*}

For a desired prior strength $S = \alpha_{k, 0} + \beta_{k, 0}$, we set:
\begin{align*}
    \alpha_{k, 0} &= S \cdot Q_\phi(o, u_k) \\
    \beta_{k, 0} &= S \cdot (1 - Q_\phi(o, u_k))
\end{align*}

This prior enforcement technique in conjunction with online learning with Thompson Sampling, as discussed in Section~\ref{subsec:thompson-sampling}, results in a stochastic policy $\pi_\theta(u_k)$ parameterized by $\theta = \left(\lbrace \left(\alpha_{k}, \beta_k\right) \rbrace_{k=1}^{K}, \phi\right)$, the learned Beta distribution shape parameters across all arms and the fixed parameters of the GQ-CNN used for initialization.

\subsection{Prior Mismatch} \label{subsec:prior-mismatch}
To measure the quality of the GQ-CNN prior, we define a notion of dissimilarity between the prior and ground truth grasp probabilities, as in \citet{empirical-TS}, termed the \textit{prior mismatch}. However, unlike \citet{empirical-TS}, which primarily focuses on mismatch between the mean of the prior distribution and true Bernoulli parameter, we present a new metric based on the discrepancy between how arms are ranked under the prior and under the ground truth distribution.

Given the grasp quality estimates of the GQ-CNN prior $q_p = \left( Q_\phi(o, u_k) \right)_{k=1}^{K}$ and the ground truth grasp probabilities $q_g = \left( p_{k}\right)_{k=1}^{K}$ on all $K$ arms, let $\mathcal{P} = \{(q_p[k], q_g[k])\}_{k=1}^{K}$. We then compute Kendall's tau coefficient, defined as:
\begin{align*}
    \tau = \frac{N_c - N_d}{  \sqrt{ (N_c + N_d + T_p)(N_c + N_d + T_g) }  }
\end{align*}
where $N_c$ and $N_d$ are the number of concordant and discordant pairs in $\mathcal{P}$, respectively, and $T_p$ and $T_g$ are the number of pairs for which $q_p[i] = q_p[j]$ and $q_g[i] = q_g[j]$, respectively~\cite{kendall1938new,kendall1945treatment}. As a rank correlation coefficient, $\tau \in [-1, 1]$, where $1$ denotes a perfect match in the rankings and $-1$ denotes perfectly inverse rankings. We define the prior mismatch $M$ as a dissimilarity measure that maps $\tau$ to $[0, 1]$:
\begin{align*}
    M = \frac{1 - \tau}{2}
\end{align*}
In practice, to control for stochasticity when sampling arms on the initial observation $o$, we average $M$ over 10 independently sampled sets of $K$ arms.

\section{Practical Implementation}
\label{subsec:practical}

\begin{figure*}[p]
    \centering
    \includegraphics[width=\linewidth]{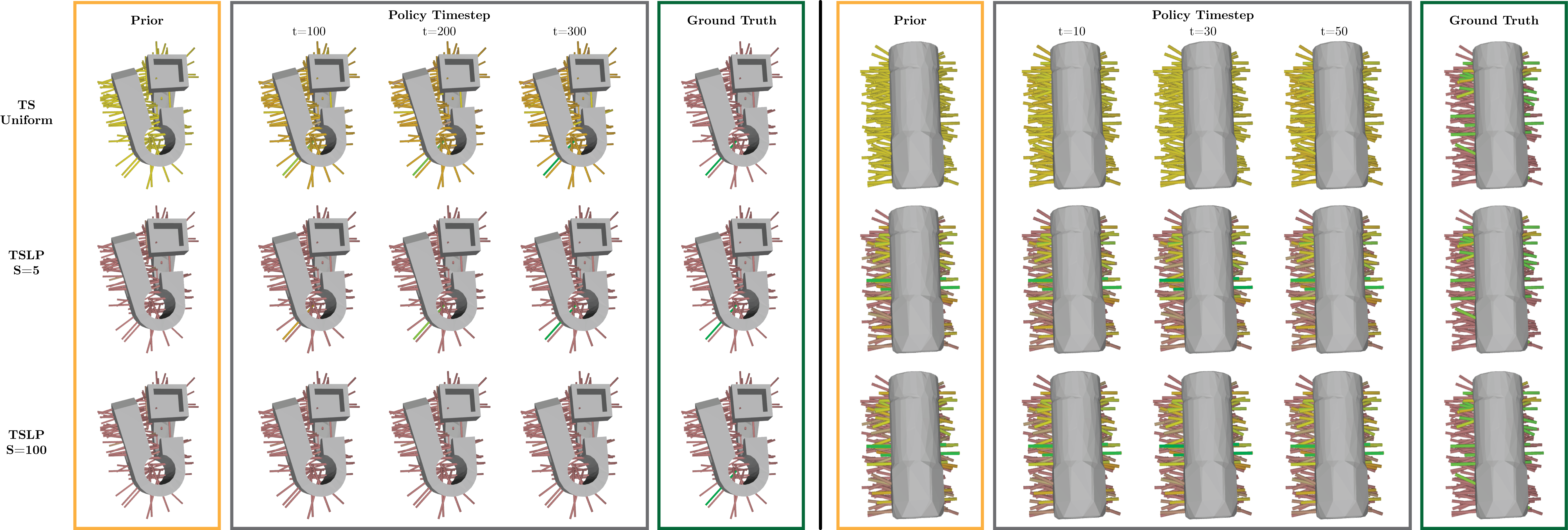}
    \caption{We visualize the evolution of the mean Bernoulli parameter (defined in Equation~\eqref{eq:est_grasp_success_prob}) inferred by \algabbr with varying prior strengths on sampled arms over learning steps for two different objects. Grasps with high estimated success probabilities or ground truth quality values are colored green, while those with low estimated success probabilities or ground truth qualities are colored orange or red. The inferred mean Bernoulli parameter for \algabbr eventually converges to the ground truth probabilities. For the first object, we note that  \algabbr is able to find the best grasps when the prior strength is relatively weak, but unable to do so when the prior strength is too high since the prior is overly pessimistic ($M = 0.46$). For the second object, the prior is relatively good ($M = 0.31$), so increasing the prior strength accelerates discovery of the best grasps.}
\label{fig:whisker-plot}
\end{figure*}

We implement the method from Section~\ref{sec:methods} in a simulated environment using 3D object models from the Dex-Net 4.0 dataset~\cite{mahler2019learning}. We render a simulated depth image of the object using camera parameters that are selected to be consistent with a Photoneo PhoXi S industrial depth camera. Arms are selected by sampling parallel-jaw antipodal grasp candidates on the observation $o$ using the antipodal image grasp sampling technique from Dex-Net 2.0 \cite{mahler2017dex}. The antipodal grasp sampler thresholds the depth image to find areas with high gradients, then uses rejection sampling over pairs of pixels to find antipodal grasp points. Each parallel jaw grasp is represented by a center point $\mathbf{p} = (x,y,z) \in \mathds{R}^3$ and a grasp axis $\mathbf{v} \in \mathds{R}^3$~\cite{mahler2016dex}. They are visualized as whiskers in Figures~\ref{fig:splash} and~\ref{fig:whisker-plot}. Once the arms are sampled from the image, we calculate the prior grasp probabilities using GQ-CNN, then deproject each grasp from image space into a 3D grasp using the known camera intrinsics. Note that \algabbr can also be easily be applied with different types of grasps such as Suction grasps~\cite{mahler2018dex} provided that the actions corresponding to the arms are parameterized accordingly. We then iteratively choose grasps according to the policy for a set number of timesteps and collect the reward for each grasp.

\begin{figure}[ht!]
  \centering
  \begin{minipage}{\linewidth}
     \begin{algorithm}[H]
     \caption{\algname for Image-Space Grasp Exploration}
     \label{alg:main}
     \begin{algorithmic}
     \renewcommand{\algorithmicrequire}{\textbf{Input:}}
     \renewcommand{\algorithmicensure}{\textbf{Output:}}
     
     \REQUIRE Number of arms ($K$), Maximum Iterations $T$, Pretrained GQ-CNN $Q_\phi(\cdot, \cdot)$, Prior Strength $S$
     \ENSURE  Grasp exploration policy: $\pi_\theta(u_k)$
     \STATE Capture observation $o$, sample $K$ antipodal grasps $\{a_k\}_{k=1}^{K}$, and compute prior beliefs $\alpha_{k, 0}, \beta_{k, 0} \ \forall \ k \in \{1, 2, \hdots K\}$ using $Q_\phi(o, u_k)$ using method from Section~\ref{subsec:NN-priors}.
      \FOR{$t=1, ..., T$}
          \STATE Select action $u_k$ using Thompson sampling as in Section~\ref{subsec:thompson-sampling}
          \STATE Execute $u_k$ and observe $r(u_k)$ 
          \STATE Update $\alpha_{k, t}, \beta_{k, t} \ \forall k \in \{1, 2, \hdots K\}$ as in Section~\ref{subsec:thompson-sampling}
    \ENDFOR
    \end{algorithmic}
    \end{algorithm}
  \end{minipage}
\end{figure}

Algorithm~\ref{alg:main} summarizes the full approach discussed in Section~\ref{sec:methods} along with implementation details. If we are unable to sample $K$ arms or if none of the corresponding grasps has ground truth quality greater than zero, we do not consider the object pose. In simulation, we evaluate the probability of grasp success for each arm using the robust wrench resistance metric, which measures the grasp's ability to resist the gravity wrench under perturbations in the grasp pose, as in \cite{mahler2018dex}. Then, rewards during policy learning and evaluation are sampled from a Bernoulli distribution with parameter defined by this metric. Note that while computing this metric requires knowledge of the object geometry, this metric is simply used to simulate grasp success on a physical robotic system and is not exposed to \algabbr.
\section{Experiments}

\begin{figure*}[t!]
    \centering
    \vspace{4pt}
    \includegraphics[width=0.85\linewidth]{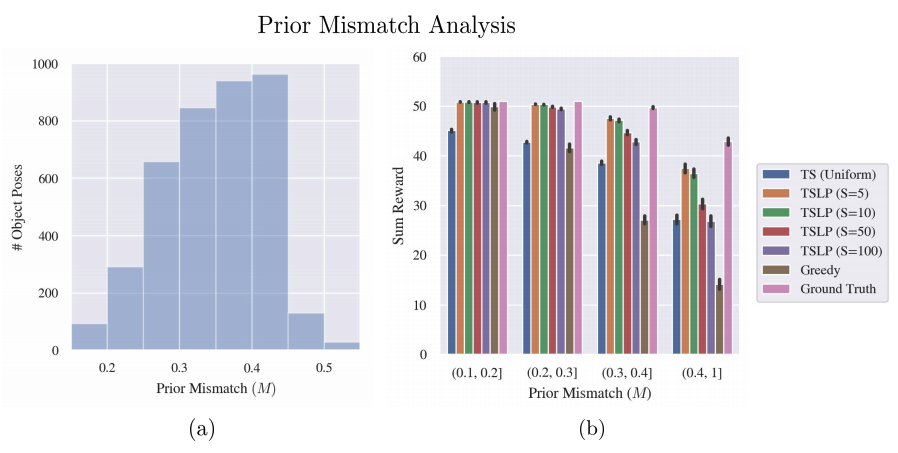}
    \caption{(a) The distribution of prior mismatch $M$ for the $3946$ total
    object poses in the dataset used by~\cite{mahler2019learning}. We find that $M$ ranges from $0.16$ to $0.64$ and a value between $0.4$ and $0.45$ is the most common, accounting for about $25\%$ of the object poses. All object poses with $M$ above $0.55$ are placed into the highest bin. (b) The sum of rewards over policy evaluation computed over $3000$ object poses randomly selected from the dataset. This plot suggests that the chosen metric accurately describes the mismatch between the prior and ground truth grasp quality distribution, as performance of all \algabbr policies decreases with increased prior mismatch.}
    \label{fig:mismatch}
\end{figure*}

\begin{figure*}[p]
    \centering
    \includegraphics[width=\linewidth]{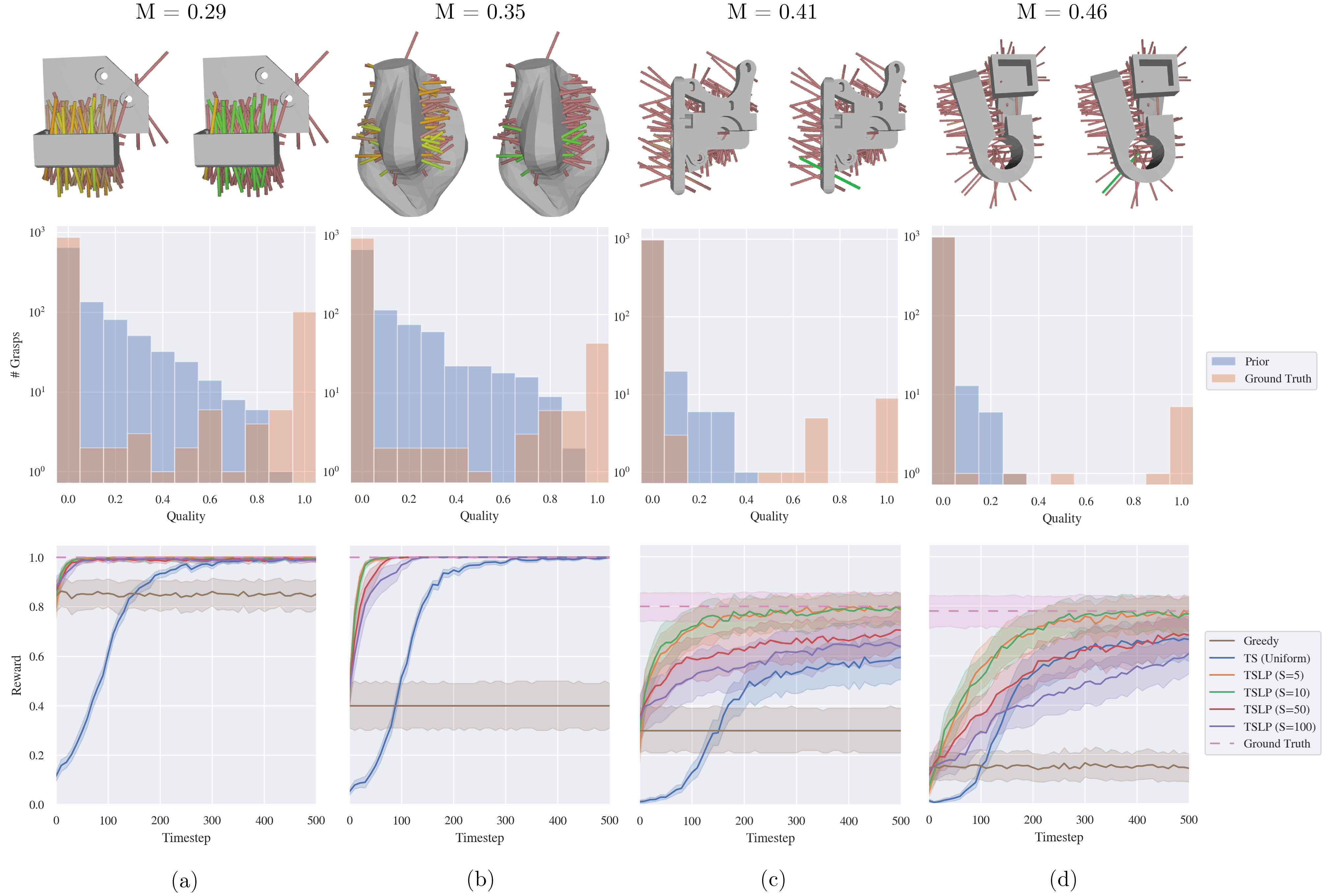}
    \caption{Visualization of policy performance for all baselines and \algabbr policies (labeled with their prior strength). The first row visualizes grasp qualities as measured by the GQ-CNN prior (left) and the ground truth grasp success probabilities (right) for a single stable pose of each of the four objects (shown top down). Green whiskers indicate high estimated or ground truth grasp quality, while red whiskers indicate low estimated or ground truth grasp quality. In the second row, we visualize the distributions of GQ-CNN prior and ground truth grasp qualities. (a) With a low prior mismatch ($M=0.29$), the greedy policy performs well and all Thompson sampling policies with non-zero prior strengths converge quickly to the ground truth. (b-c) For objects with higher prior mismatch, the Thompson sampling policies with non-zero prior strength rapidly improve on the prior for object poses with higher prior mismatch ($M=0.35$ and $M=0.40$, respectively). (d) For objects with very high prior mismatch ($M=0.46$), the Thompson sampling policies with non-zero prior strength converge more slowly, but still show improvement on the baseline with prior strength $0$.}
    \label{fig:policy_learning_curves}
\end{figure*}

\begin{table*}[ht!]
\caption {\textbf{Policy evaluation on large object set:} We evaluate each \algabbr policy, the greedy and Thompson sampling with uniform prior baselines, and the ground truth policy on a dataset of $3000$ object poses and report the average sum reward over all runs on each of the object poses ($300,000$ total training runs per policy, $100$ training runs per object for each policy) in the format of mean $\pm$ standard deviation. Since we evaluate the policy $51$ times per episode, the maximum possible sum reward is $51$. For readability, we scale all results by a factor of $100/51$ for a maximum scaled sum reward of $100$. We find that the best performing \algabbr policy, \algabbr (S=5), outperforms the greedy baseline by $64.5\%$ while achieving performance within $5.7\%$ of the ground truth oracle baseline.}
\label{tab:all-rollouts}
\begin{center}
\begin{tabu} to 0.9\linewidth {X[c]X[c]X[c]X[c]X[c]X[c]X[c]}
\toprule
 \textbf{Greedy} & \textbf{TS (Uniform)} & \textbf{\algabbr(S=5)} & \textbf{\algabbr (S=10)} & \textbf{\algabbr (S=50)} & \textbf{\algabbr (S=100)} & \textbf{Ground Truth} \\
\midrule
$54.33 \pm 33.02$ & $72.08 \pm 20.67$ & $\mathbf{89.37 \pm 17.88}$ & $88.43 \pm 18.53$ & $82.63 \pm 23.51$ & $78.88 \pm 26.29$ & $94.53 \pm 13.49$ \\
\bottomrule
\end{tabu}
\end{center}
\end{table*}

\subsection{Setup}
In simulation experiments, we evaluate both the accuracy of the prior mismatch metric and the ability of \algabbr to increase grasp exploration efficiency. We assess whether \algabbr can discover higher quality grasps than baselines which do not explore online or which explore online but do not leverage learned priors for the grasp selection policy. In both experiments, we make use of the dataset from \citet{mahler2019learning}, which contains approximately 1,600 object meshes. 

We evaluate the learned policies every 10 steps of learning, and perform 500 learning steps in total for all experiments. To evaluate the learned policies, we sample 100 grasps from the current policy without policy updates and compute the metric defined in Equation~\eqref{eq:objective}. We evaluate \algabbr with a variety of different prior strengths to evaluate how important the GQ-CNN prior is for policy performance. We also compare to Thompson sampling with a uniform prior over the arms. Thus, this policy does not utilize the GQ-CNN prior at all, and all learning is performed online. Note that when evaluating policies, there are two key sources of uncertainty: (1) the variability in the arms sampled on the initial observation $o$, and (2) the inherent stochasticity during learning given a set of arms. To control for variations in these parameters, when reporting results on a particular pose of an object, $10$ different sets of $K = 100$ arms are sampled on the corresponding observation $o$. Then, for each of these sets of arms, every policy is trained $10$ times for a total of $100$ rollouts for each object pose.

We additionally compare the learned policy to a greedy policy that repeatedly selects the grasp with highest quality under GQ-CNN as in~\cite{mahler2019learning} and a ground truth oracle policy, which repeatedly selects the grasp with the highest quality under the ground truth grasp quality metrics computed in simulation. The former gives an idea of policy performance if no online exploration is performed, while the latter provides an upper bound on possible performance since it can access the true grasp success probabilities, which are not available to our algorithm.

\subsection{Simulation Experiments}
We conduct simulation experiments across object poses with a wide range of prior mismatches $M$, as shown in Figure~\ref{fig:mismatch}(a), which plots the frequency of prior mismatch values over the $3946$ total object poses in the dataset. When the prior mismatch is relatively low, we expect policies which give more weight to the prior to perform well, while if the prior mismatch is high, we expect policies which prioritize online exploration over following the prior to attain higher rewards.

We evaluate each policy on $3000$ of these object poses and compute the sum reward of all policies averaged over the $300,000$ total training runs ($100$ training runs per object pose). The results are shown in Figure~\ref{fig:mismatch}(b) and Table~\ref{tab:all-rollouts}. Figure~\ref{fig:mismatch}(b) shows policy performance as a function of prior mismatch, given the distribution of objects over prior mismatch values shown in Figure~\ref{fig:mismatch}(a) over $3000$ total object poses. These results suggest that the metric introduced here accurately models prior mismatch, as increased prior mismatch causes performance for all online learning policies, as well as the greedy policy, to degrade. A second trend is that object poses with higher prior mismatch also tend to have lower ground truth quality values, suggesting that GQ-CNN especially struggles to identify high quality grasps when very few are present or when the highest quality grasps have comparatively lower quality.

Table~\ref{tab:all-rollouts} shows that \algabbr significantly outperforms the greedy baseline and is able to achieve average total reward that is very close to the ground truth policy. This result suggests that \algabbr is able to successfully leverage priors from GQ-CNN to outperform GQ-CNN on a wide variety of objects of varying geometries.

As a further case study, we select a set of 4 objects, as shown in Figure~\ref{fig:policy_learning_curves}, which are diverse in their shapes and sizes and vary widely in their prior mismatch $M$. As expected, the ground truth policy (GT) achieves the best performance since it uses oracle information. We find that for objects with relatively low prior mismatch ($M=0.29$), the greedy policy and the Thompson sampling policies which place very high weight on the GQ-CNN prior (high prior strength) perform very well. However, for objects with higher prior mismatch ($M=0.35$, $M = 0.41$), we find that the greedy policy performs much more poorly, and online exploration is critical to finding high quality grasps. However, even with high prior mismatch, the gap in performance between the Thompson sampling policies that use the prior and the uniform prior Thompson sampling policy indicates that the GQ-CNN prior helps accelerate grasp exploration substantially. Finally, for objects with very high prior mismatch ($M = 0.46$), the greedy policy and Thompson sampling policies with high prior strengths perform poorly, as expected. However, Thompson sampling policies with low prior strength outperform Thomspon sampling with a uniform prior. This result indicates that although the prior is of very low quality, it still provides useful guidance to the Thompson sampling policy if a low prior strength is used.

Figure~\ref{fig:whisker-plot} shows how the mean Bernoulli parameter inferred by \algabbr evolves over learning steps for each of the sampled arms. \algabbr is able to successfully learn grasp qualities close to the ground truth grasp qualities for a wide variety of different objects. Note that the learned policy is generally more accurate for higher quality grasps, which makes sense since Thompson sampling directs exploration towards high reward grasps, allowing it to focus on distinguishing between high quality grasps rather than capturing the quality distribution of low quality grasps. For the first object, \algabbr is able to find the best grasp when the prior strength is relatively weak, but performs poorly when the prior strength is set too high. For the second object, the prior mismatch is lower, so increasing the prior strength accelerates discovery of the best grasps on the object. Note that with a uniform prior, Thompson sampling is generally able to discover most of the best grasps, but fails to distinguish them from bad grasps, resulting in poorer policy performance when these grasps are sampled during policy evaluation.
\section{Discussion and Conclusion}

In this paper, we present \algname, a bandit exploration strategy for robotic grasping which facilitates use of expressive neural network-based prior belief distributions and enables efficient online exploration for objects for which this prior is inaccurate. We quantify the notion of prior mismatch as it pertains to the ranking of arms and explore the effect of prior strength on the efficiency and efficacy of online learning. Experiments suggest that across a dataset of 3000 object poses, \algabbr outperforms both a greedy baseline as well as a Thompson sampling baseline that uses a uniform prior and is able to leverage a GQ-CNN prior to significantly accelerate grasp exploration.

\section{Future Work}
\label{sec:future-work}
In future work, we will design new online learning algorithms to explore grasps across different object stable poses and extend experiments to new grasping modalities, such as suction. In addition, we will explore ways to approximately enforce pose consistency in physical experiments. For example, we can use a string to lift the object after each grasp and put it into pose. Additionally, we can detect stable pose changes by evaluating whether the observed depth image changes in a way that cannot be described by a planar rotation and translation. Using the Super4PCS algorithm \cite{Super4PCSLibrary}, we can compute the registration of the new point cloud with respect to the original point cloud and restrict the range of output to planar transformations. If the algorithm cannot find such a planar transformation, we resample grasps on the new pose.

\section*{Acknowledgments}
\footnotesize
\noindent This research was performed at the AUTOLAB at UC Berkeley in affiliation with the Berkeley AI Research (BAIR) Lab, Berkeley Deep Drive (BDD), the Real-Time Intelligent Secure Execution (RISE) Lab, and the CITRIS "People and Robots" (CPAR) Initiative. Authors were also supported by the Scalable Collaborative Human-Robot Learning (SCHooL) Project, a NSF National Robotics Initiative Award 1734633, and in part by donations from Google and Toyota Research Institute. Ashwin Balakrishna and Michael Danielczuk are supported by the National Science Foundation Graduate Research Fellowship Program under Grant No. 1752814. This article solely reflects the opinions and conclusions of its authors and do not reflect the views of the sponsors. We thank our colleagues who provided feedback and suggestions, in particular Brijen Thananjeyan.

\renewcommand*{\bibfont}{\footnotesize}
\printbibliography 

@STRING{icra = {{Proc. {IEEE} Int. Conf. Robotics and Automation (ICRA)}}}

@STRING{iros = {Proc. IEEE/RSJ Int. Conf. on Intelligent Robots and Systems (IROS)}}

@STRING{case = {Proc. {IEEE} Conf. on Automation Science and Engineering (CASE)}}

@STRING{cvpr = {Proc. {IEEE} Conf. on Computer Vision  and Pattern Recognition (CVPR)}}

@STRING{ijrr = {Int. Journal of Robotics Research (IJRR)}}

@STRING{rss = {Proc. Robotics: Science and Systems (RSS)}}

@STRING{corl = {Conf. on Robot Learning (CoRL)}}

@inproceedings{johns2016deep,
  title={Deep learning a grasp function for grasping under gripper pose uncertainty},
  author={Johns, Edward and Leutenegger, Stefan and Davison, Andrew J},
  booktitle=iros,
  pages={4461--4468},
  year={2016},
  organization={IEEE}
}

@incollection{empirical-TS,
    title = {An Empirical Evaluation of Thompson Sampling},
    author = {Olivier Chapelle and Li, Lihong},
    booktitle = {Advances in Neural Information Processing Systems 24},
    editor = {J. Shawe-Taylor and R. S. Zemel and P. L. Bartlett and F. Pereira and K. Q. Weinberger},
    pages = {2249--2257},
    year = {2011},
    publisher = {Curran Associates, Inc.},
    url = {http://papers.nips.cc/paper/4321-an-empirical-evaluation-of-thompson-sampling.pdf}
}

@inproceedings{laskey2015multi,
  title={Multi-armed bandit models for 2d grasp planning with uncertainty},
  author={Laskey, Michael and Mahler, Jeff and McCarthy, Zoe and Pokorny, Florian T and Patil, Sachin and Van Den Berg, Jur and Kragic, Danica and Abbeel, Pieter and Goldberg, Ken},
  booktitle={2015 IEEE International Conference on Automation Science and Engineering (CASE)},
  pages={572--579},
  year={2015},
  organization={IEEE}
}

@inproceedings{mahler2016dex,
  title={Dex-net 1.0: A cloud-based network of 3d objects for robust grasp planning using a multi-armed bandit model with correlated rewards},
  author={Mahler, Jeffrey and Pokorny, Florian T and Hou, Brian and Roderick, Melrose and Laskey, Michael and Aubry, Mathieu and Kohlhoff, Kai and Kr{\"o}ger, Torsten and Kuffner, James and Goldberg, Ken},
  booktitle={2016 IEEE International Conference on Robotics and Automation (ICRA)},
  pages={1957--1964},
  year={2016},
  organization={IEEE}
}

@article{TS-survey,
  author = {Russo, Daniel J. and Roy, Benjamin Van and Kazerouni, Abbas and Osband, Ian and Wen, Zheng},
  title = {now publishers - A Tutorial on Thompson Sampling},
  number = 1,
  pages = {1--96},
  publisher = {Now Publishers},
  volume = 11,
  year = 2018
}

@inproceedings{mahler2017dex,
  title={Dex-net 2.0: Deep learning to plan robust grasps with synthetic point clouds and analytic grasp metrics},
  author={Mahler, Jeffrey and Liang, Jacky and Niyaz, Sherdil and Laskey, Michael and Doan, Richard and Liu, Xinyu and Ojea, Juan Aparicio and Goldberg, Ken},
  booktitle=rss,
  year={2018}
}

@inproceedings{mahler2018dex,
  title={Dex-Net 3.0: Computing Robust Robot Vacuum Suction Grasp Targets in Point Clouds using a New Analytic Model and Deep Learning},
  author={Mahler, Jeffrey and Matl, Matthew and Liu, Xinyu and Li, Albert and Gealy, David and Goldberg, Ken},
  booktitle=icra,
  year={2018}
}

@inproceedings{wang2019adversarial,
  title={Adversarial Grasp Objects},
  author={Wang, David and Tseng, David and Li, Pusong and Jiang, Yiding and Guo, Menglong and Danielczuk, Michael and Mahler, Jeffrey and Ichnowski, Jeffrey and Goldberg, Ken},
  booktitle={2019 IEEE 15th International Conference on Automation Science and Engineering (CASE)},
  pages={241--248},
  year={2019},
  organization={IEEE}
}

@incollection{prattichizzo2016grasping,
  title={Grasping},
  author={Prattichizzo, Domenico and Trinkle, Jeffrey C},
  booktitle={Springer handbook of robotics},
  pages={955--988},
  year={2016},
  publisher={Springer}
}

@book{murray2017mathematical,
  title={A mathematical introduction to robotic manipulation},
  author={Murray, Richard M},
  year={2017},
  publisher={CRC press}
}

@inproceedings{pinto2016supersizing,
  title={Supersizing self-supervision: Learning to grasp from 50k tries and 700 robot hours},
  author={Pinto, Lerrel and Gupta, Abhinav},
  booktitle=icra,
  pages={3406--3413},
  year={2016},
  organization={IEEE}
}

@inproceedings{kalashnikov2018qt,
  title={Qt-opt: Scalable deep reinforcement learning for vision-based robotic manipulation},
  author={Kalashnikov, Dmitry and Irpan, Alex and Pastor, Peter and Ibarz, Julian and Herzog, Alexander and Jang, Eric and Quillen, Deirdre and Holly, Ethan and Kalakrishnan, Mrinal and Vanhoucke, Vincent and others},
  booktitle=corl,
  year={2018}
}

@inproceedings{james2019sim,
  title={Sim-to-real via sim-to-sim: Data-efficient robotic grasping via randomized-to-canonical adaptation networks},
  author={James, Stephen and Wohlhart, Paul and Kalakrishnan, Mrinal and Kalashnikov, Dmitry and Irpan, Alex and Ibarz, Julian and Levine, Sergey and Hadsell, Raia and Bousmalis, Konstantinos},
  booktitle=cvpr,
  pages={12627--12637},
  year={2019}
}

@article{mahler2019learning,
  title={Learning ambidextrous robot grasping policies},
  author={Mahler, Jeffrey and Matl, Matthew and Satish, Vishal and Danielczuk, Michael and DeRose, Bill and McKinley, Stephen and Goldberg, Ken},
  journal={Science Robotics},
  volume={4},
  number={26},
  pages={eaau4984},
  year={2019},
  publisher={Science Robotics}
}

@article{bohg2013data,
  title={Data-driven grasp synthesis—a survey},
  author={Bohg, Jeannette and Morales, Antonio and Asfour, Tamim and Kragic, Danica},
  journal={IEEE Transactions on Robotics},
  volume={30},
  number={2},
  pages={289--309},
  year={2013},
  publisher={IEEE}
}

@article{levine2018learning,
  title={Learning hand-eye coordination for robotic grasping with deep learning and large-scale data collection},
  author={Levine, Sergey and Pastor, Peter and Krizhevsky, Alex and Ibarz, Julian and Quillen, Deirdre},
  journal=ijrr,
  volume={37},
  number={4-5},
  pages={421--436},
  year={2018},
  publisher={SAGE Publications Sage UK: London, England}
}

@inproceedings{bousmalis2018using,
  title={Using simulation and domain adaptation to improve efficiency of deep robotic grasping},
  author={Bousmalis, Konstantinos and Irpan, Alex and Wohlhart, Paul and Bai, Yunfei and Kelcey, Matthew and Kalakrishnan, Mrinal and Downs, Laura and Ibarz, Julian and Pastor, Peter and Konolige, Kurt and others},
  booktitle=icra,
  pages={4243--4250},
  year={2018},
  organization={IEEE}
}

@article{viereck2017learning,
  title={Learning a visuomotor controller for real world robotic grasping using simulated depth images},
  author={Viereck, Ulrich and Pas, Andreas ten and Saenko, Kate and Platt, Robert},
  journal={arXiv preprint arXiv:1706.04652},
  year={2017}
}

@inproceedings{morrison2018closing,
	title={{Closing the Loop for Robotic Grasping: A Real-time, Generative Grasp Synthesis Approach}},
	author={Morrison, Douglas and Corke, Peter and Leitner, J\"urgen},
	booktitle={Proc.\ of Robotics: Science and Systems (RSS)},
	year={2018}
}

@article{lenz2015deep,
  title={Deep learning for detecting robotic grasps},
  author={Lenz, Ian and Lee, Honglak and Saxena, Ashutosh},
  journal=ijrr,
  volume={34},
  number={4-5},
  pages={705--724},
  year={2015},
  publisher={SAGE Publications Sage UK: London, England}
}

@article{saxena2008robotic,
  title={Robotic grasping of novel objects using vision},
  author={Saxena, Ashutosh and Driemeyer, Justin and Ng, Andrew Y},
  journal=ijrr,
  volume={27},
  number={2},
  pages={157--173},
  year={2008},
  publisher={Sage Publications Sage UK: London, England}
}

@inproceedings{kappler2015leveraging,
  title={Leveraging big data for grasp planning},
  author={Kappler, Daniel and Bohg, Jeannette and Schaal, Stefan},
  booktitle=icra,
  pages={4304--4311},
  year={2015},
  organization={IEEE}
}

@inproceedings{bicchi2000robotic,
  title={Robotic grasping and contact: A review},
  author={Bicchi, Antonio and Kumar, Vijay},
  booktitle=icra,
  volume={1},
  pages={348--353},
  year={2000},
  organization={IEEE}
}

@book{rimon2019mechanics,
  title={The Mechanics of Robot Grasping},
  author={Rimon, Elon and Burdick, Joel},
  year={2019},
  publisher={Cambridge University Press}
}

@article{kendall1945treatment,
  title={The treatment of ties in ranking problems},
  author={Kendall, Maurice G},
  journal={Biometrika},
  volume={33},
  number={3},
  pages={239--251},
  year={1945},
  publisher={JSTOR}
}

@article{kendall1938new,
  title={A new measure of rank correlation},
  author={Kendall, Maurice G},
  journal={Biometrika},
  volume={30},
  number={1/2},
  pages={81--93},
  year={1938},
  publisher={JSTOR}
}

@misc {Super4PCSLibrary,
    author = {Mellado, Nicolas and Aiger, Dror and Mitra, Niloy J.},
    title = {Super 4PCS Library},
    howpublished = {https://github.com/nmellado/Super4PCS},
    year = {2017},
}

@inproceedings{pedersen:hal-02495837,
  TITLE = {{Grasping Unknown Objects by Coupling Deep Reinforcement Learning, Generative Adversarial Networks, and Visual Servoing}},
  AUTHOR = {Pedersen, Ole-Magnus and Misimi, Ekrem and Chaumette, Fran{\c c}ois},
  URL = {https://hal.inria.fr/hal-02495837},
  BOOKTITLE = {{ICRA 2020 -  IEEE International Conference on Robotics and Automation}},
  ADDRESS = {Paris, France},
  PUBLISHER = {{IEEE}},
  PAGES = {1-8},
  YEAR = {2020},
  MONTH = May,
  HAL_ID = {hal-02495837},
  HAL_VERSION = {v1},
}

@article{Bodnar2019QuantileQF,
  title={Quantile QT-Opt for Risk-Aware Vision-Based Robotic Grasping},
  author={Cristian Bodnar and Adrian Li and Karol Hausman and Peter Pastor and Mrinal Kalakrishnan},
  journal={ArXiv},
  year={2019},
  volume={abs/1910.02787}
}

@article{Liu2020DeepDG,
  title={Deep Differentiable Grasp Planner for High-DOF Grippers},
  author={Min Liu and Zherong Pan and Kai Xu and Kanishka Ganguly and Dinesh Manocha},
  journal={ArXiv},
  year={2020},
  volume={abs/2002.01530}
}


\end{document}